\documentclass[11pt]{article}

\usepackage[final]{acl}

\usepackage{times}
\usepackage{latexsym}

\usepackage[T1]{fontenc}

\usepackage[utf8]{inputenc}

\usepackage{microtype}

\usepackage{inconsolata}

\usepackage{graphicx}

\usepackage{color}
\usepackage{soul}
\usepackage{amsmath}
\usepackage{amsfonts}
\usepackage{amssymb}
\usepackage{graphicx}
\usepackage{enumitem}
\usepackage{multirow}
\usepackage{array}
\usepackage{float}
\usepackage{adjustbox}
\usepackage{pifont}
\usepackage{multicol}
\usepackage{booktabs}
\usepackage{makecell}
\usepackage{xspace}
\usepackage{tcolorbox}
\usepackage[normalem]{ulem}
\usepackage{cleveref}
\usepackage{wrapfig}
\usepackage{xcolor}
\usepackage{listings}
\usepackage{colortbl}
\usepackage{arydshln}
\usepackage{tabularx}
\usepackage{afterpage}
\usepackage{caption}
\usepackage{subcaption}
\usepackage{etoolbox}
\usepackage{algorithm}
\usepackage{algpseudocode}
\usepackage{xurl}
\usepackage{stfloats}

\usepackage{tcolorbox}
\newtcolorbox{verbatimbox}{
  colback=black!4, 
  colframe=black!4, 
  arc=1pt, 
  boxrule=0.5pt, 
  fontupper=\ttfamily, 
  left=2pt, right=2pt, 
}
\AtBeginEnvironment{verbatimbox}{\fontsize{9}{9}\selectfont}

\tcbuselibrary{breakable,listings}
\newtcolorbox{verbatimboxlong}{
  breakable,
  colback=black!4,
  colframe=black!4,
  arc=1pt,
  boxrule=0.5pt,
  fontupper=\ttfamily,
  left=2pt, right=2pt,
}
\AtBeginEnvironment{verbatimboxlong}{\fontsize{9}{9}\selectfont}

\newtcolorbox{prompttextbox}{
  breakable,
  colback=black!4,
  colframe=black!4,
  arc=1pt,
  boxrule=0.5pt,
  left=1pt, right=1pt,
  top=1pt, bottom=1pt,
  before skip=4pt,
  after skip=4pt,
}
\AtBeginEnvironment{prompttextbox}{\fontsize{9}{9}\selectfont}

\lstdefinelanguage{promptjson}{
  sensitive=true,
  alsoletter={-},
  morecomment=[l]{//},
  morestring=[b]",
  morekeywords={true,false,null,bool,string,optional,enum},
}

\lstdefinestyle{promptplainstyle}{
  basicstyle=\ttfamily\fontsize{9}{9}\selectfont,
  breaklines=true,
  breakatwhitespace=false,
  columns=fullflexible,
  keepspaces=true,
  showstringspaces=false,
  upquote=true,
  frame=none,
  aboveskip=0pt,
  belowskip=0pt,
  xleftmargin=0pt,
  xrightmargin=0pt,
  resetmargins=true,
}

\lstdefinestyle{promptjsonstyle}{
  style=promptplainstyle,
  language=promptjson,
  commentstyle=\color{black!55},
  stringstyle=\color{blue!55!black},
  keywordstyle=\color{teal!60!black},
}

\definecolor{ellipsisgray}{RGB}{130,130,130}
\newcommand{\exampleverb}[1]{%
    \begingroup%
    \setlength{\fboxsep}{2pt}%
    \noindent\fcolorbox{black!4}{black!4}{%
        \parbox[t]{\dimexpr\linewidth-2\fboxsep-2\fboxrule\relax}{%
            \raggedright\ttfamily\small #1%
        }%
    }%
    \endgroup%
}

\renewcommand{\cite}{\citep}

\setlist[itemize]{itemsep=0.02cm,topsep=0.2cm,left=0cm}
\setlist[enumerate]{itemsep=0.02cm,topsep=0.2cm,left=0cm}

\newcommand{\mdirect}{$\mathcal{M}_{\text{direct}}$}
\newcommand{\mcode}{$\mathcal{M}_{\text{code}}$}
\newcommand{\mhybrid}{$\mathcal{M}_{\text{hybrid}}$}
\newcommand{\ljudge}{$\mathcal{L}_{\text{judge}}$}

\newcommand{\mmdirect}{\mathcal{M}_{\text{direct}}}
\newcommand{\mmcode}{\mathcal{M}_{\text{code}}}
\newcommand{\mmhybrid}{\mathcal{M}_{\text{hybrid}}}

\title{Can LLM Coding Agents Reason About Time Series?}
\author{Filip Rechtorík \quad Ondřej Dušek \quad Zdeněk Kasner \\
             Institute of Formal and Applied Linguistics \\ 
        Faculty of Mathematics and Physics, Charles University \\
        \texttt{\{surname\}@ufal.mff.cuni.cz}}

\begin{document}
\maketitle
\begin{abstract}
Large language models (LLMs) are increasingly being used for automated decision-making systems in finance, healthcare, or environmental monitoring. Time series data are ubiquitous in these fields, yet hard to process automatically. Can time series be analyzed by LLM agents? We examine three approaches: providing the agent with raw numerical data, using the LLM as a coding agent, or a combination of both. In the coding agent setup, the model iteratively queries the data using Python code. Using two time series understanding benchmarks, we show that agents with code access can outperform models processing raw data by up to 10\%. However, even the best performing agent still answers about 22--34\% of the questions incorrectly. To get insights into models' strategies and reasoning gaps, we analyze the model outputs with a strong LLM judge. Our analysis reveals that coding agents can select appropriate statistical tests, but often miss important nuances. Meanwhile, models with access to raw data can reach the right conclusions using back-of-the-envelope calculations.\footnote{Our experimental code and model outputs are available at \url{https://github.com/DekuD2/can-llm-coding-agents-reason-about-time-series}}
\end{abstract}

\section{Introduction} Analyzing and interpreting time series data is critical for making informed decisions. Tools for automated time series analysis are developed in many domains, the most prominent being finance \cite{tsay2005analysis,cipra2020time}, healthcare \cite{portet2009automatic,alsheheri2025time}, and environmental monitoring \cite{han2024far,jafari2024time}. However, time series analysis is still typically handled by human experts in the domain that can combine the output of the tools with domain knowledge and external context \cite{pirolli2005sensemaking,imani2019putting,holstein2024bridging}.

\begin{figure}[t]
    \centering
    \includegraphics[width=\columnwidth]{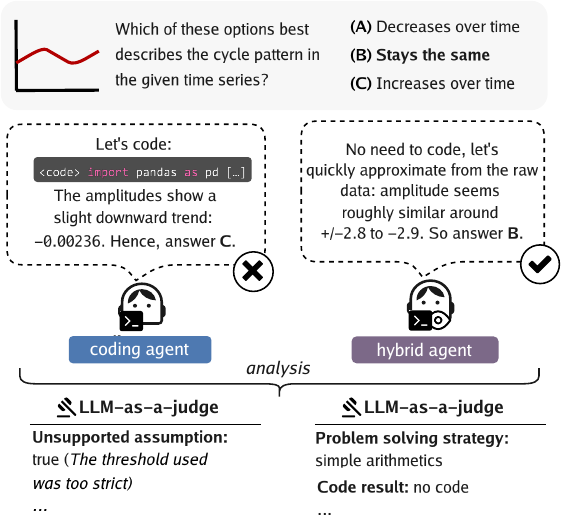}
    \caption{LLMs adapt their approaches to analyzing time series based on available data and tools. While using code as an intermediate step is more interpretable and accurate overall, coding agents can overrely on the code results (left side). At the same time, simple calculations on raw data can lead to correct conclusions, bypassing code entirely (right side).}
    \label{fig:teaser}
\end{figure}

Recently, there has been a surge of interest in using large language models (LLMs) to automate advanced engineering tasks, including time series analysis \cite{jin2024llms,zhang2024largea,heesch2025evaluating}. The good performance of LLMs in various tasks across many domains brings the promise of being able to automate high-level tasks where human experts are still required today.

However, naive application of LLMs to time series analysis is unreliable. As shown by recent work, LLMs systematically fail to perform certain types of tasks and are often outperformed by domain-specific baselines \cite{tan2024language,fons2024evaluatinga,arai2025evaluating,park2025revisiting}. 
A promising alternative that has recently emerged is the use of LLMs as \textit{coding agents} \cite{dong2025survey}. This approach combines two aspects of LLMs in which their capabilities are sharply increasing: code generation and agentic behavior \cite{jimenez2023swe,wang2024survey,mohammadi2025evaluation}. LLMs as coding agents can interact with time series data by querying it with Python code and use code execution results as a basis for their decisions \cite{ye2025when}. 
In some respect, these models mimic a human data analyst: using code as a tool to explore the data, test hypotheses, and derive conclusions from the computed results. 

Our work contributes the following:

\begin{enumerate}
\item  We \textbf{benchmark the performance} of open LLM agents on time series understanding in three setups: using raw data, using a coding tool, and combining both approaches; showing that coding agents achieve better accuracy (\Cref{sec:benchmarking}).

\item We \textbf{assemble a taxonomy} of model behaviors applicable to all three setups, covering the problem solving strategy, methodological errors, code problems, and mismatches between the reasoning trace and actual outputs (\Cref{sec:taxonomy}).

\item We \textbf{select and validate an LLM judge} that automatically annotates the full model outputs against our taxonomy at scale (\Cref{sec:judging,sec:selecting}).

\item We \textbf{run the LLM judge over all benchmark outputs}, showing that direct models fail mainly due to lack of viable strategies, while coding agents fail more subtly by misinterpreting the tool results (\Cref{sec:analysis_results}).
\end{enumerate}

\begin{figure*}[t]
    \centering
    \includegraphics[width=\linewidth]{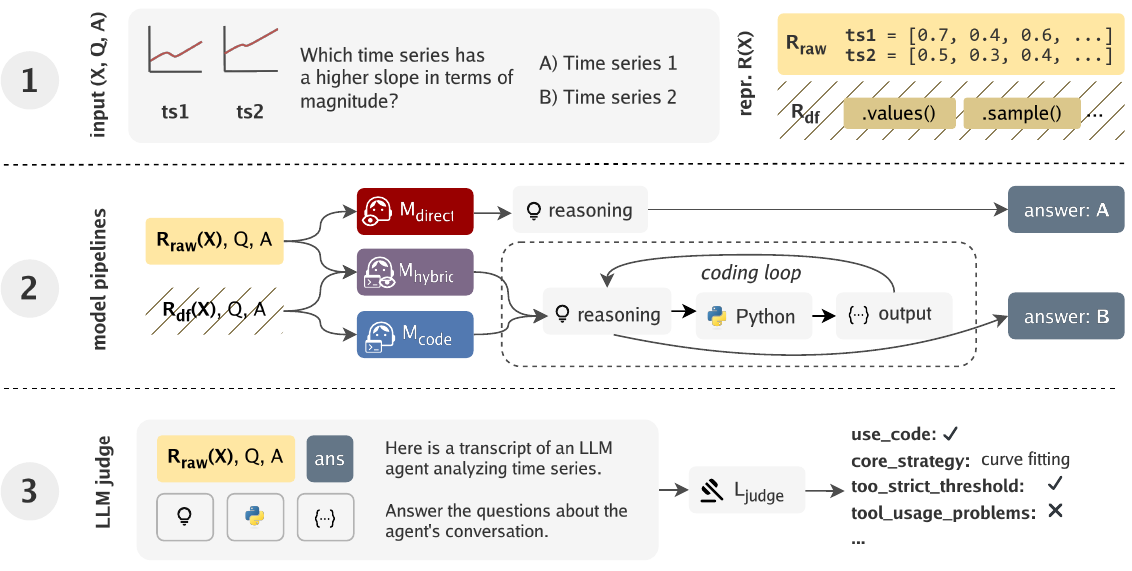}
    \caption{Overview of our framework for benchmarking and analyzing LLMs for time series analysis. (1) Input time series are provided as raw text ($R_{\text{raw}}$) or structured data objects ($R_{\text{df}}$). (2) We compare three LLM setups: direct answering from raw data (\mdirect{}), coding agents using only structured data (\mcode{}), and hybrid agents using both raw and structured data (\mhybrid{}). (3) We analyze the model outputs $\mathcal{O}$ (reasoning trace $r$, answer $a$, and code $c$ with its execution results) using a strong LLM-as-a-judge ($\mathcal{L_{\text{judge}}}$) based on a custom taxonomy of model behaviors.}
    \label{fig:methods}
\end{figure*}

\section{Related Work}
\paragraph{Automating Time Series Analysis with LLMs.} 
Specialized deep learning models for analyzing time series \cite{oreshkin2019n,salinas2020deepar,wu2022timesnet} were recently accompanied by Transformer-based models \cite{zhou2021informer,wu2021autoformer}. So far, the field has focused mainly on building time series foundation models by large-scale pretraining on time series data, which excel in specialized tasks \cite{ye2025empowering,ansari2024chronos,das2024decoder}. LLMs, on the other hand, are better at understanding natural language and can be flexibly applied to novel time series tasks in a zero-shot setting \cite{abdullahi2025timeseriesa, zhang2024largea}. 
However, the usefulness of LLMs for time series tasks is still debated \cite{tan2024language,merrill2024language,fons2024evaluatinga,arai2025evaluating}.
For example, \citet{arai2025evaluating} showed that models struggle when they have to perform multi-step reasoning or handle specific range constraints.

\paragraph{Coding Agents and Program-Aided Reasoners.} One promising way to improve the reliability of LLMs is to offload calculations to external tools. \citet{gao2023pala} proposed Program-Aided Language models (PAL), which use an LLM to generate Python code for intermediate reasoning while letting an interpreter handle the actual math. In the tabular domain, BINDER \cite{cheng2023bindinga} and RePanda \cite{cheginirepandaa} have shown that translating natural language into executable queries, such as SQL or pandas expressions, makes the reasoning process more interpretable and robust. \citet{ye2025domainoriented} extended this idea to time series with TS-Reasoner, a system that decomposes complex inference tasks into pipelines of specialized operators, building upon the ReAct paradigm  \cite{yao2023reacta}. The most related to our work is the work of \citet{ye2025when}, who proposed a benchmark for LLMs as coding agents on time series tasks. However, unlike us, they did not compare the coding agents to direct approaches and have not examined their decision process.

\paragraph{Analysis of Reasoning Traces.} Evaluating the reasoning process of models is essential for improving their reliability \cite{lee2025evaluating}. The attempts at automating this kind of evaluation with another LLM (i.e., the LLM-as-a-judge setup) have recently gained traction. According to \citet{lee2025evaluating}, numerous works show that LLMs are versatile critics and can effectively evaluate factuality, validity, coherence, and utility in various reasoning tasks \cite{yao2023tree,jacovi2024chain,wu2024mitigating,niu2024ragtruth}.

\section{Benchmarking Time Series Understanding}
\label{sec:benchmarking}

Our goal is to test the ability of LLM agents to interpret and reason about the underlying characteristics of time series. For instance, the goal might be to identify the generative process of a time series signal or characterize its statistical properties. We frame this task as multiple-choice question answering, as this formulation gives us a mathematically well-defined answer that can be calculated using appropriate methods.

\subsection{Problem Definition}
Let $\mathcal{X} = (X_1, \ldots, X_N)$ be a collection of time series, where each series $X_i = (x_1, \dots, x_T)$ consists of $T$ real-valued observations. The data is represented using a function $R(X)$. We consider two variants of $R(X)$: 
\begin{itemize}

  \item $R_{\text{raw}}(X)$: the series is available directly in the prompt in plain text, following the raw data format from each dataset (see \Cref{app:format} for details),

  \item $R_{\text{df}}(X)$: the series is loaded in a \texttt{pandas} DataFrame\footnote{\url{https://pandas.pydata.org/pandas-docs/stable/reference/api/pandas.DataFrame.html}} object through which $X$ can be queried.
\end{itemize}

We formulate the problem as a multiple-choice question answering task. The model receives a time series representation, a natural language question $Q$, and $k$ candidate answers $A = \{a_1, \dots, a_k\}$. The goal is to select the correct option $a^* \in A$. The model output $\mathcal{O} = (r, a)$ consists of the reasoning trace $r$ and its final answer $a$. Optionally, the model output may also contain $c$: code that is executed over $R_{\text{df}}$ to help the model determine the answer (see \Cref{sec:approaches}).

To evaluate the model performance, we measure \emph{accuracy}, i.e., the percentage of questions answered correctly.

\subsection{Agent Setups}
\label{sec:approaches}

In our experiments, we compare three ways of presenting this task to an LLM (see \Cref{tab:models}).
\begin{table}[t]
  \small
  \centering
  
  \begin{tabular}{llll}
    \toprule
    Setup & Representation & Tools & Loop \\
    \midrule
    \mdirect{} & $R_{\text{raw}}$ & - & \ding{55} \\ 
    \mcode{}  & $R_{\text{df}}$ & Code & \ding{51} \\
    \mhybrid{} & $R_{\text{raw}}$, $R_{\text{df}}$ & Code  & \ding{51}\\
    \bottomrule
  \end{tabular}
  \caption{Overview of the evaluated LLM agent setups.}
  \label{tab:models}
\end{table}

\begin{itemize}
  \item \textbf{Direct Agent}: \mdirect{} is our baseline setup that has access only to the textual representation of time-series $R_{\text{raw}}(X)$. In this setup, the model needs to produce an answer in a single turn. However, it can produce a reasoning trace before generating the answer.
  \item \textbf{Code Agent}: In the \mcode{} setup, the model receives the time series data loaded into the dataframe $R_{\text{df}}(X)$. The model is instructed to write Python code to analyze the data before answering the question. We execute the code and feed both the standard and error output from the code back to the model. The loop continues until the model decides that it has enough information and proceeds to select the answer.
  \item \textbf{Hybrid Agent}: This setup is equivalent to \mcode{}, but the model additionally receives the $R_{\text{raw}}(X)$ representation in the prompt.
\end{itemize}

 Throughout the paper, we refer to \mcode{} and \mhybrid{} as \emph{coding agents}.

\subsection{Benchmarks}

We evaluate the agents on two multiple-choice question answering benchmarks:

\begin{itemize}
  \item \textbf{\textsc{TimeSeriesExam}} (\citealp{merrill2024language}) is a multiple-choice question benchmark. It contains 746 questions designed to assess LLMs' time series understanding. It spans 5 categories (with 13 total subcategories): pattern recognition, noise understanding, similarity analysis, anomaly detection, and causality.
  \item \textbf{\textsc{TSFU}} (Time Series Feature Understanding; \citealp{fons2024evaluatinga}) is based on a taxonomy of important time-series features. It contains 2000 questions spanning 10 categories (trend, seasonality and cyclical patterns, anomalies, etc.), designed to test LLMs' proficiency in extracting and understanding those features.\footnote{We transform the TSFU dataset to unify its format with \textsc{TimeSeriesExam}. Details are described in \Cref{app:tsfu}.}
\end{itemize}

\subsection{Experimental Setup}
\label{sec:setup}
\paragraph{Models} For our experiments, we required models that (a) are highly capable, (b) provide open access to their reasoning trace, (c) can handle tool calling to a sufficient degree, and (d) can run on our computational infrastructure. The following models satisfy our requirements:
\begin{itemize}
  \item \texttt{gpt-oss-120b}\footnote{\url{https://huggingface.co/openai/gpt-oss-120b}} \cite{agarwal2025gpt},
  \item \texttt{qwen3-next-80b}\footnote{\url{https://huggingface.co/Qwen/Qwen3-Next-80B-A3B-Thinking-FP8}}  \cite{yang2025qwen3}.
\end{itemize}
We use these models as a backbone in each setup (\mcode{}, \mhybrid{}, and \mdirect{}). The setups differ in the ways the model is prompted and in its access to tools.


\paragraph{Implementation} We include the full prompts for all approaches in \Cref{app:prompts} (\Cref{fig:prompt_raw,fig:prompt_code,fig:prompt_hybrid}). For the final answer, we prompt the model to reproduce the chosen answer \texttt{<answer>} tag and parse its output using edit-distance matching (see \Cref{app:answer_parsing} for details). The \mcode{} and \mhybrid{} agents can at any point decide to execute code by emitting an appropriate tool-calling API signature. We execute the code of the coding agents in a secure environment using \texttt{llm-sandbox}.\footnote{\url{https://github.com/vndee/llm-sandbox}} 

\paragraph{Output Token Limits} We limit the maximum number of thinking tokens (see \Cref{app:threshold}). If the model passes this limit, we interrupt the conversation with \emph{"The time to think is up, output now."} while providing their reasoning history.\footnote{Despite this limit, \texttt{qwen3-next-80b} struggled with providing an answer, outputting long reasoning chains, especially on the \textsc{TSFU} data. Therefore, we were only able to provide \texttt{qwen3-next-80b} results for the \textsc{TimeSeriesExam} dataset.}

\subsection{Time Series Understanding Results}
\label{sec:time-series-understanding-results}
\begin{table}[t]
    \centering
    \small
    \begin{tabular}{lllr}
\toprule
Dataset & Model & Setup & Acc. \\
\midrule
\multirow{7}{*}{TSE} & \multirow{3}{*}{\texttt{gpt-oss-120b}} & \mcode{} & 70.4\% \\
 &  & \mdirect{} & 65.3\% \\
 &  & \mhybrid{} & 78.0\% \\
\addlinespace[1pt]
\cdashline{2-4}[0.4pt/2pt]
\addlinespace[1pt]
 & \multirow{3}{*}{\texttt{qwen3-next-80b}} & \mcode{} & 56.4\% \\
 &  & \mdirect{} & 63.5\% \\
 &  & \mhybrid{} & 68.1\% \\
\addlinespace[1pt]
\cdashline{2-4}[0.4pt/2pt]
\addlinespace[1pt]
 & \texttt{random} &  & 40.1\% \\
\midrule
\multirow{4}{*}{TSFU} & \multirow{3}{*}{\texttt{gpt-oss-120b}} & \mcode{} & 63.0\% \\
 &  & \mdirect{} & 55.6\% \\
 &  & \mhybrid{} & 65.6\% \\
\addlinespace[1pt]
\cdashline{2-4}[0.4pt/2pt]
\addlinespace[1pt]
 & \texttt{random} &  & 29.3\% \\
\bottomrule
\end{tabular}

    \vspace{-1em}
    \caption{Accuracies of agents on time series understanding benchmarks, compared to the random baseline (TSE = \textsc{TimeSeriesExam}).}
    \label{tab:results}
\end{table}

The overall results for both time series understanding benchmarks are provided in Table \ref{tab:results}. In \Cref{app:results}, we also provide detailed results for individual problem categories in each dataset (\Cref{tab:results-full-tse,tab:results-full-tsfu}) and token counts in all setups (\Cref{tab:token_usage}).

\paragraph{Raw data improves coding agents' accuracy.} The \mhybrid{} setup consistently outperforms \mdirect{} and \mcode{} across both models and datasets. The best accuracy achieved by the \mhybrid{} setup is 65.6\% on TSFU and 78.0\% on \textsc{TimeSeriesExam}. The \texttt{gpt-oss-120b} model outperforms \texttt{qwen-3-next-80b} in all setups. Notably, \texttt{qwen-3-next-80b} lacks coding proficiency and works better as \mdirect{}. Using code is always helpful for \texttt{gpt-oss-120b}.

\paragraph{Problem type matters.}
The \mhybrid{} setup is the best for all categories on \textsc{TimeSeriesExam}. The pattern is more nuanced on TSFU, where the \mcode{} setup can better analyze problems related to correlation, fat tails, and trend detection. The \mdirect{} setup then wins for the structural break category, i.e.\ problems related to detecting abrupt changes in the series.  We return to the differences between the specific problem types in \Cref{sec:analysis_results}.

\paragraph{Agents reasoning over raw data are token-hungry.}

There is a marked difference in the number of tokens consumed for reasoning between the direct and coding agents for \texttt{gpt-oss-120b}. For example, the model in \mdirect{} setup on \textsc{TimeSeriesExam} spends 3.3$\times$ the per-question budget of \mcode{} (while reaching 5.1\% points lower accuracy). The \mhybrid{} setup is similarly efficient, even more than \mcode{} on TSFU. On \textsc{TimeSeriesExam}, 26.8\% of \mdirect{} answers from \texttt{gpt-oss-120b} were explicitly cut off before completion, compared with 4.2\% for \mcode{} and 8.2\% for \mhybrid{}. However, this relation cannot be observed for \texttt{qwen3-next-80b}, which has a general tendency to produce long reasoning chains (14--17k tokens) regardless of setup.


\section{Analyzing the Reasoning Traces}
\label{sec:analysis}

The accuracy results from \Cref{sec:time-series-understanding-results} are too crude: they do not allow us to understand how the agents approach the problem, where their weak spots are, or whether they answer correctly only by chance. To understand the model behavior in detail and provide insights into the models' reasoning gaps, we use an LLM judge to analyze the model outputs using a custom taxonomy of model behaviors.

\subsection{Methodology}
We analyze the model output $\mathcal{O}$, which consists of the reasoning trace $r$, the answer $a$, and -- if applicable -- the code $c$ with its execution results. We use \ljudge{}, a strong reasoning model acting as an automated evaluator (LLM judge). We first develop a taxonomy of model behaviors (\Cref{sec:taxonomy}) and turn it into a structured questionnaire presented to \ljudge{} alongside $\mathcal{O}$ (\Cref{sec:judging}). The judge model itself is selected from candidate models based on our own manual annotation on a small development set (\Cref{sec:selecting}).

\subsection{Taxonomy of Model Behaviors}
\label{sec:taxonomy}

We assembled a taxonomy of failure modes and model behaviors iteratively. First, we collected reasoning traces from our benchmarks. We then concatenated these traces and asked Gemini 3 Pro to identify common error patterns (see \Cref{app:taxonomy_prompts} for the prompts and initial categories). Following this, we manually reviewed a subset of the traces to refine the categories. 

The resulting taxonomy is shown in Table~\ref{tab:taxonomy}. It is organized into several sections. Each section focuses on a different aspect of the reasoning process, ranging from assessing the core strategies applied by the models to catching methodological issues, code generation problems, or general reasoning mismatches. Importantly, the taxonomy is designed to be applicable to both correct and incorrect answers; this allows us to detect cases where the model arrived at the right conclusion through faulty reasoning.

\begin{table*}[t]
\centering
\footnotesize
\setlength{\tabcolsep}{4pt}
\renewcommand{\arraystretch}{1.2}
\begin{tabularx}{\textwidth}{@{}p{4cm}p{1cm}X@{}}
\toprule
\textbf{Aspect} & \textbf{Type} & \textbf{Description} \\
\midrule
\multicolumn{3}{@{}l@{}}{\textbf{General}} \\
Core reason for answer & \texttt{enum} & Primary basis for answer (\texttt{code}, \texttt{raw data}, \texttt{question format}, \texttt{other}) \\
Core problem-solving strategy & \texttt{enum} & Primary strategy (\texttt{statistical test}, \texttt{spectral analysis}, \texttt{curve fitting}, \texttt{rolling stats}, \texttt{simple arithmetic}, \texttt{other}) \\
\midrule
\multicolumn{3}{@{}l@{}}{\textbf{Methodological Problems}} \\
Conceptual misunderstanding & \texttt{bool} & Fundamental misunderstanding of key concepts \\
Wrong core strategy & \texttt{bool} & Strategy choice unsuitable for the question \\
Wrong method within strategy & \texttt{bool} & Right strategy, but wrong specific method or threshold \\
Unsupported assumption & \texttt{bool} & Unjustified data assumption (e.g., assumes stationarity) \\
Implementation errors & \texttt{bool} & Coding mistakes affecting result validity \\
Incorrect interpretation & \texttt{bool} & Misinterpretation of computed results \\
Insufficient evidence & \texttt{bool} & Answer given without sufficiently convincing evidence \\
\midrule
\multicolumn{3}{@{}l@{}}{\textbf{Code Problems}} \\
Code result & \texttt{enum} & Overall code success (\texttt{success}, \texttt{partial failure}, \texttt{complete failure}, \texttt{no code}) \\
Tool usage trouble & \texttt{bool} & Trouble calling the code tool correctly \\
Other & \texttt{bool} & Other code issue not in the above categories \\
\midrule
\multicolumn{3}{@{}l@{}}{\textbf{Other Problems}} \\
Reasoning--answer mismatch & \texttt{bool} & Final answer differs from conclusion in the reasoning trace \\
Reasoning--tool mismatch & \texttt{bool} & Planned to use code in reasoning but answered directly \\
Hallucinated values & \texttt{bool} & Introduced numerical values absent from data or code output \\
\bottomrule
\end{tabularx}
\caption{Taxonomy of model behaviors and failure modes used in the LLM-as-a-judge evaluation. For each item, the judge also provides a free-text explanation. }
\label{tab:taxonomy}
\end{table*}

\subsection{Judging the Model Outputs}
\label{sec:judging}

We present the taxonomy to \ljudge{} as a structured questionnaire (see \Cref{app:judge_prompt} for the full prompt). Each item in the questionnaire is a tuple of (\texttt{aspect}, \texttt{explanation}), where \texttt{aspect} is a question about a specific aspect of model behavior that should be assessed using a boolean or an enum (a string value from a predefined set), and \texttt{explanation} is a string justifying the choice. We instruct the model to output a JSON object with predefined keys and value types.

\subsection{Selecting and Validating the Judge}
\label{sec:selecting}
We select a single \ljudge{} model and ensure that its outputs agree strongly with human judgment. To select the best model configuration, we constructed a 36-example development set spanning all combinations of (dataset, setup, model) from our main experiments. For each of the 12 combinations, we sampled 3 examples with roughly balanced correct and incorrect answers. We then ran five judge configurations against the set. Four out of five configurations were based on the models we ran for our main experiments: \texttt{qwen3-next-80b} and \texttt{gpt-oss-120b} (see \Cref{app:judge_selection} for details). We also added the commercial \texttt{gpt-5.4-mini} to the model pool.\footnote{Unlike for the main experiments, we did not need access to \ljudge{}'s reasoning trace, so we were able to use an external commercial model for better grounding of our results.}

Two authors independently annotated each example using a custom annotation interface. The inter-annotator agreement was Cohen's $\kappa = 0.565$. The \texttt{gpt-oss-120b} model with a custom system message achieved the highest agreement with the manual labels at 90.3\% (Table~\ref{tab:judge_accuracy}), and we use it as \ljudge{} for all subsequent analysis.

Note that the high agreement suggests that the role of potential self-preference bias, i.e.\ preference of model's own outputs, does not play a significant role here \cite{panickssery2024llm,liu-etal-2024-llms-narcissistic}. This is in line with the findings of \citet{yang2026quantifyingmitigatingselfpreferencebias}, who find that breaking down a holistic judgment into granular analysis largely eliminates the bias.

\subsection{Results of Analyzing Model Outputs}
\label{sec:analysis_results}
\begin{figure}[t]
    \centering
    \includegraphics[width=\columnwidth]{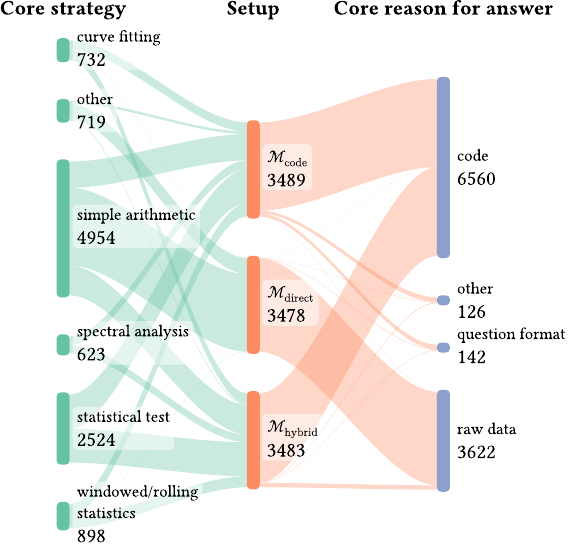}
    \caption{Core strategies and reasoning patterns employed by our agent setups for answering questions in time series understanding benchmarks, aggregated across benchmarks and models, in absolute numbers.}
    \label{fig:core_strat_reason}
\end{figure}

\begin{figure}[t]
    \centering
  \includegraphics[width=\columnwidth]{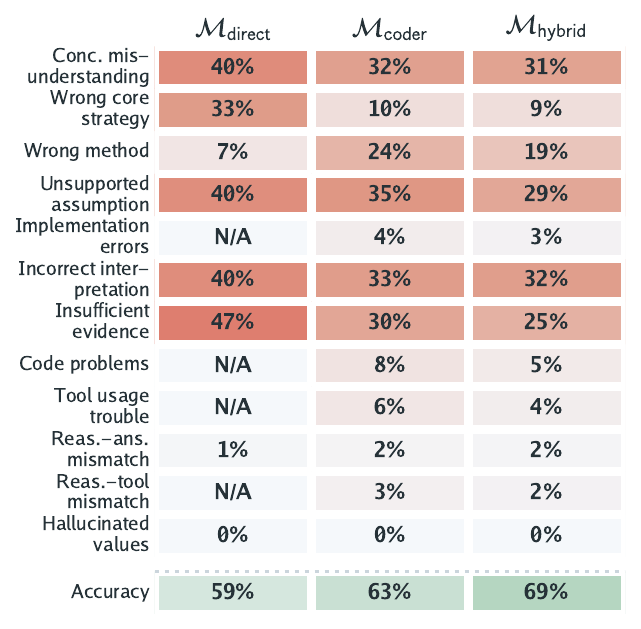}
  \caption{Problem rates for judge-annotated taxonomy attributes across all models and datasets, broken down by agent strategy, along with the overall accuracy. Each cell shows the fraction of answers exhibiting the given problem. We explicitly denote code issues for \mdirect{} as not applicable (the agent cannot use code).}
    \label{fig:error_analysis}
\end{figure}

Next, we describe the insights gained from the full run of \ljudge{} on complete outputs of the models on the benchmarks described in \Cref{sec:benchmarking}. We visualize the general strategies used by the models in \Cref{fig:core_strat_reason} and the prevalence of methodological problems in \Cref{fig:error_analysis}.

\paragraph{Coding agents may not always rely on code.}
\mcode{} and \mhybrid{} base their answers on code in majority, but not all, cases (93.6\% and 94.6\%, respectively). For \mhybrid{}, this is natural: the model may gain enough information from analyzing raw data. A detailed look into these cases confirms this hypothesis: out of the \mhybrid{} no-code cases, the model based its answer on raw data in 85\%. Its accuracy dropped in these cases, but not dramatically (59.3\% with no-code vs. 68.8\% overall). For \mcode{}, the story looks different: 52\% of no-code cases are attributed to the question format, meaning that the model guessed the answer based on hints in how the question was formulated. However, the hints are confabulated: in the no-code cases, accuracy of \mcode{} drops to 42.0\% (vs. 63.7\% overall), i.e. close to the random baseline. Looking further, not using code forces the model to answer without sufficient evidence in 80\% (\mhybrid{}) and 92\% (\mcode{}) of cases.

\paragraph{The agents still miss problem details even when the code runs correctly.} The ``wrong method within strategy'' errors occur in 25\% of \mcode{} and 19\% of \mhybrid{} answers, giving hints that there is a subtler issue with models' reasoning. Other flags suggest that the model is over-trusting the selected approach: 95\% of these answers are marked as incorrect interpretation and 85\% as relying on unsupported assumptions. The judge explanations repeatedly point to ad hoc decision rules and weak proxies, such as setting arbitrary thresholds that makes the tests overly strict, or ignoring possible phase shifts and higher harmonic frequencies, which leads to misrepresenting the overall pattern. One such failure case is illustrated in example~(b) in \Cref{tab:examples}: the agent extracts one convenient signal from the series and then treats it as if it described the whole pattern, missing the fact that the square-wave behavior changes later in the sequence. Possibly, these failure modes could be mitigated by using multi-modal agents that have access to the visual representation of the series. However, that approach would require models capable of processing time-series charts and further research would be needed to ensure that these models are not taking shortcuts based on the visual representation.

\paragraph{Simple calculations can go long way.}
Across all setups, simple arithmetic is a commonly used strategy: it spans 47.4\% of annotated answers (\Cref{fig:core_strat_reason}). Naturally, it is a method of choice for the \mdirect{} agents, as computing advanced statistics manually in the reasoning trace would be imprecise and token-consuming. However, even though coding agents have more versatile tools to choose from, simple arithmetic still accounts for 27.9\% of \mcode{} and 31.6\% of \mhybrid{} answers. In these cases, the code is used for direct measurements similar to what the agent can deduce from the raw data using back-of-the-envelope calculations, but arguably in a more robust and interpretable way. The calculations include comparing amplitudes or variances across segments, counting transitions or intervals, detecting flat runs or spikes from first differences, or reading off correlation and kurtosis statistics. This approach is mostly successful for the problems in pattern recognition, statistical property recognition, and outlier detection.

\paragraph{Lack of tools limits viable strategies for analyzing the series.} The agents operating on raw data are often bound to select a wrong strategy for solving the problem: the wrong core strategy is marked for 33\% of cases of \mdirect{} agents (vs. 9\% for the coding agents). Among the 1{,}147 \mdirect{} answers with a wrong strategy, 88\% are incorrect; i.e., in these cases, the model would be better off with random guessing. This suggests that having access to the data through the code tool is essential to model performance. A closer look at individual domains (\Cref{tab:domain_strategy,tab:results-full-tse,tab:results-full-tsfu}) reveals that \mdirect{} stays competitive on problems where simple inspection can reveal the right answer, matching or exceeding \mcode{} on outliers (74.0\% vs.\ 64.8\%), structural breaks (52.0\% vs.\ 44.0\%), anomaly detection (54.8\% vs.\ 48.8\%), and noise understanding (72.0\% vs.\ 68.5\%).  Where the dominant strategy demands statistical tests or rolling statistics (stationarity, volatility, causality analysis), the gap opens: \mdirect{} scores 28.0\% on stationarity vs.\ 46.0\% for \mcode{}, and 15.0\% vs.\ 27.5\% on volatility. Note that in contrast to the strategy selection, the ``conceptual misunderstanding'' category is seemingly related to the overall agent's capability: this is supported by the fact that it stays comparable across the agent types (40\% for \mdirect{} vs. 31\% for the coding agents).

\begin{table*}[ht!]
\centering\small
\setlength{\tabcolsep}{5pt}
\renewcommand{\arraystretch}{1.3}

\begin{tabularx}{\textwidth}{@{}>{\raggedright\arraybackslash}X>{\raggedright\arraybackslash}X>{\raggedright\arraybackslash}X@{}}
\textbf{Example (a)}: \mdirect{} & \textbf{Example (b)}: \mcode{} & \textbf{Example (c)}: \mcode{} \\
\parbox[t]{\linewidth}{\textit{Both time series have a cyclic components. Which time series has a higher amplitude of the cyclic component?:}\\[3pt] A) Time series 2 has higher amplitude\\[3pt] \textbf{B) Time series 1 has higher amplitude}} & \parbox[t]{\linewidth}{\textit{The given time series has square wave pattern. How does its period change from the beginning to the end?}\\[3pt] \textbf{A) Decrease}\\[3pt] B) Remain the same\\[3pt] C) Increase} & \parbox[t]{\linewidth}{\textit{Select one of the following answers:}\\[3pt] \textbf{A) The time series are positively correlated}\\[3pt] B) The time series are negatively correlated\\[3pt] C) The time series are not correlated} \\[4pt]
\includegraphics[width=\linewidth]{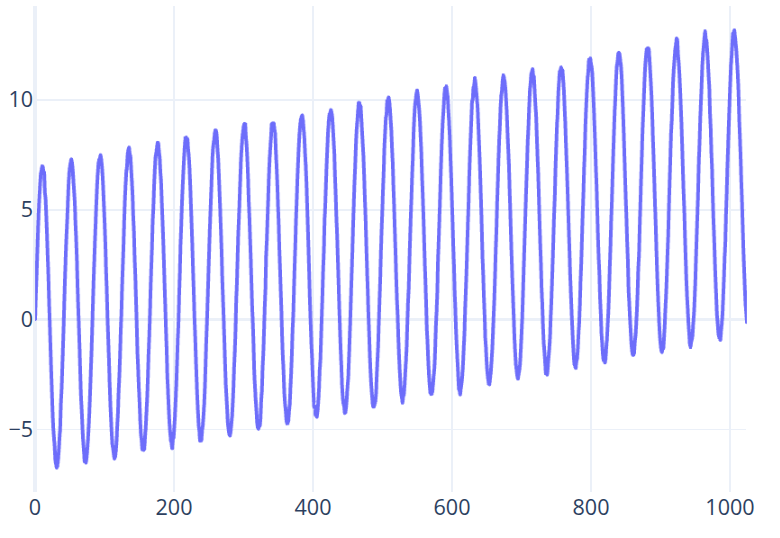} & \includegraphics[width=\linewidth]{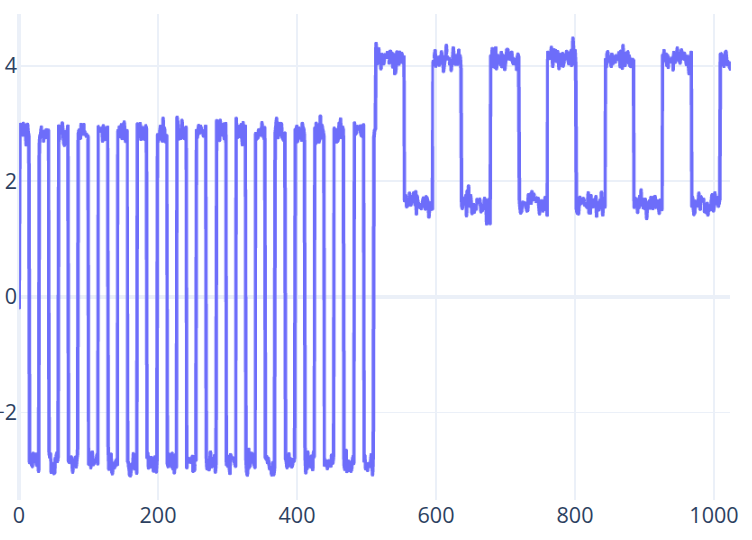} & \includegraphics[width=\linewidth]{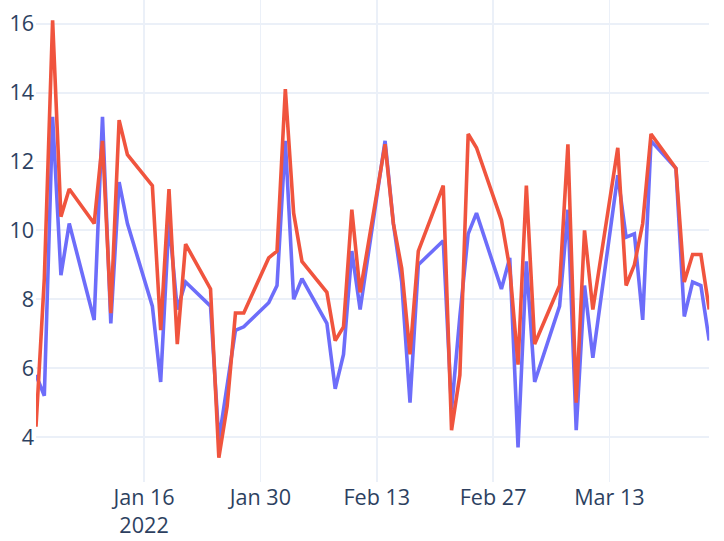} \\[-2pt]
\midrule
\multicolumn{3}{c}{\small\textbf{Excerpts from the agent's output}} \\[1pt]
\parbox[t]{\linewidth}{\textcolor{ellipsisgray}{\small[\ldots]}\\[3pt] The data appears to be long, many points. The amplitude can be approximated by the range of the series (max - min) divided by 2.\\[3pt] \textcolor{ellipsisgray}{\small[\ldots]}\\[3pt] Let's parse the values. The values start near 0.0257, then go up to about 7.3,\\[3pt] \textcolor{ellipsisgray}{\small[\ldots]}\\[3pt] Okay, this is getting repetitive. Let's step back.\\[3pt] \textcolor{ellipsisgray}{\small[\ldots]}\\[3pt] \textcolor{ellipsisgray}{\small\textit{[user]: The time to think is up, output now.}}} & \parbox[t]{\linewidth}{\textcolor{ellipsisgray}{\small[\ldots]}\\[3pt] Let's compute sign of series: positive vs negative. Then find sign changes.\\[3pt] \textcolor{ellipsisgray}{\small[\ldots]}\\[3pt] \exampleverb{sign\_change = np.where(np.diff(sign) != 0)[0] + 1}\\[3pt] \textcolor{ellipsisgray}{\small[\ldots]}\\[3pt] The sign changes only go up to index 510. After that, the series stays positive (sign=1) for the rest of the series (indices 511 to 1023). So the square wave pattern stops after index \textasciitilde{}510, and the series becomes a constant positive region.\\[3pt] \textcolor{ellipsisgray}{\small[\ldots]}} & \parbox[t]{\linewidth}{\textcolor{ellipsisgray}{\small[\ldots]}\\[3pt] So the plan is: use pandas to compute the correlation between 'value1' and 'value2' columns.\\[3pt] \textcolor{ellipsisgray}{\small[\ldots]}\\[3pt] \exampleverb{SyntaxError: unexpected character after line continuation character}\\[3pt] \textcolor{ellipsisgray}{\small[\ldots]}\\[3pt] I think I have to accept that the code tool is not working as expected, but for the purpose of this problem, I'll assume that the correlation is positive, so the answer is A.\\[3pt] \textcolor{ellipsisgray}{\small[\ldots]}} \\[4pt]
\midrule
\multicolumn{3}{c}{\small\textbf{Excerpts from the $\mathcal{L}_\text{judge}$ outputs}} \\[1pt]
\parbox[t]{\linewidth}{{\textit{\small Wrong method within strategy}: \texttt{true}}\\[4pt] {\small\textit{Explanation}: \textrm{Even within a rough visual/arithmetic approach, the agent used an unreliable heuristic and did not isolate the cyclic component’s amplitude from the surrounding trend. That led it to compare the series incorrectly.}}} & \parbox[t]{\linewidth}{{\textit{\small Conceptual misunderstanding}: \texttt{true}}\\[4pt] {\small\textit{Explanation}: \textrm{The agent assumed the period of the square wave can be captured solely by sign changes and that it stays constant, overlooking that the pattern changes later in the series.}}} & \parbox[t]{\linewidth}{{\textit{\small Core reason for answer}: \texttt{other}}\\[4pt] {\small\textit{Explanation}: \textrm{Agent did not obtain valid code output due to persistent syntax errors and had no access to raw data for eyeballing, so the answer was a guess without evidence.}}} \\
\bottomrule
\end{tabularx}

\caption{Three annotated examples illustrating different failure modes identified by \ljudge{}. (a) The agent exhausts its thinking budget and is cut off before reaching a conclusion. (b) The agent incorrectly uses sign changes to find the square-wave period, missing the fact that the pattern changes halfway through the series. (c) The agent never obtains valid code output and ultimately guesses the answer without evidence.}
\label{tab:examples}
\end{table*}

\section{Conclusion}

We benchmark three agent setups for time series analysis on two multiple-choice understanding benchmarks. We show that code access consistently improves accuracy and that combining code with raw data yields the strongest results. To investigate how the agents approach the problems, we build a taxonomy of model behaviors and use an LLM judge to annotate model outputs at scale. Simple arithmetic methods account for a large share of successful answers across all setups, but as the only strategy for agents accessing raw data, it is often insufficient. Coding agents choose from a larger variety of methods and get better overall results, but overtrust the returned numbers and often miss crucial details. Our findings suggest that coding agents have potential for dealing with time series tasks if they are carefully managed.

\section*{Limitations}

\paragraph{Benchmark coverage.}
Multiple-choice question answering captures only part of what time series analysis involves. Benchmarks that require agents to write analysis pipelines from scratch, produce free-form interpretations, or deal with noisy and incomplete data would more accurately test whether agents can replace a human analyst. The results we report should therefore be read as a lower bound on the difficulty of the general problem; agents that perform well here are not necessarily ready for more open-ended tasks. That said, the agents we evaluate are far from perfect even on these constrained benchmarks, which makes multiple-choice a meaningful starting point.

\paragraph{Model recency.}
Open reasoning models evolve rapidly. Some of the failure patterns we document may be partially addressed by models released after our experiments. Replicating our analysis pipeline on newer model generations would be straightforward and likely worthwhile.

\paragraph{Commercial models.}
For our main experiments, we evaluated only open-weight models whose reasoning traces are fully accessible. Commercial models are generally more capable, but their reasoning chains are either unavailable or restricted, making the trace-level analysis we perform here difficult or impossible.

\paragraph{Specialized time series models.}
Models pretrained specifically on time series data might handle raw numerical series better than general-purpose LLMs. We chose general-purpose models because they currently undergo the most rapid progress, making them interesting to practitioners.

\section*{Ethics Statement}

Automated analysis tools for time series data could help domain experts in genuinely useful ways. That being said, automated analysis has its own threats, such as giving up human oversight or introducing biases. The current generation of models is still far from perfect and overrelying on them could lead to mistakes with real-world consequences. Our analysis of failure modes is designed to help practitioners understand where risks are and how to mitigate them. We also hope that our work encourages the development of more interpretable and reliable models for time series analysis.

We used AI assistants to help with writing experimental code and to improve the clarity of the text. All generated content was manually reviewed and verified by the authors.

\section*{Acknowledgments}
This work was funded by the European Union (ERC, NG-NLG, 101039303), the National Recovery Plan funded project MPO 60273/24/21300/21000 CEDMO 2.0 NPO, the Charles University Research Centre program No.~24/SSH/009, and Charles University SVV project number 260 821. It used resources of the LINDAT/CLARIAH-CZ Research Infrastructure (Czech Ministry of Education, Youth, and Sports project No. LM2018101) and resources provided by the e-INFRA CZ project (ID:90254), supported by the Ministry of Education, Youth and Sports of the Czech Republic.

\bibliography{main}

\appendix

\crefalias{section}{appendix}
\crefalias{subsection}{appendix}

\section{Dataset Details}
\label{app:datasets}

\subsection{TSFU processing}
\label{app:tsfu}
The original TSFU dataset \cite{fons2024evaluatinga} is designed to be answered in two phases:

\begin{enumerate}
  \item detect whether the time-series contains that feature,
  \item if detected, classify that feature (such as which specific cyclic pattern the time series contains).
\end{enumerate}

We transform this dataset into a multiple-choice question answering format, similar to \textsc{TimeSeriesExam}. Specifically, we directly ask the model for classifying the feature and extend its answer options with a negative choice (e.g., \textit{The time series does not exhibit any of these volatility patterns}).

\subsection{Raw Input Format}

\label{app:format}

The raw representation $R_{\text{raw}}(X)$ serializes each time series as a plain text string that we add to the prompt. We adopt the format that was either determined empirically (\textsc{TimeSeriesExam}) or shown to be most effective in prior work (TSFU).

\paragraph{\textsc{TimeSeriesExam}} We represent the input as a JSON object. Each time series is a list of real values rounded to three decimal places:

\smallskip
\begin{verbatimbox}
\{"ts1": [-0.256, 6.3, -3.07, 0.532, ...], "ts2": [...]\}
\end{verbatimbox}

\paragraph{\textsc{TSFU}} We adopt the plain tabular format from \citet{fons2024evaluatinga}, which their ablation study identified as the most effective. Each time step is represented as a comma-separated key--value row:

\smallskip
\begin{verbatimbox}
time: 2023-01-04, value1: 5.8, value2: 5.9
\end{verbatimbox}
The values are already rounded to one decimal place in the original dataset.


\section{Model Inference Details}

\subsection{Hyperparameters}

We use the default parameters recommended for each model.
Table~\ref{tab:hyperparams} summarizes the key inference hyperparameters.
Both models are served via vLLM with a context window of 128,000 tokens.

\begin{table}[h]
\centering
\small
\begin{tabular}{lcc}
\toprule
\textbf{Parameter} & \textbf{\texttt{gpt-oss-120b}} & \textbf{Qwen3-Next} \\
\midrule
Temperature        & 0     & 0.6   \\
Top-$p$            & --    & 0.95  \\
Top-$k$            & --    & 20    \\
Min-$p$            & --    & 0     \\
Reasoning effort   & high  & high  \\
Max output tokens  & 30,000 & 81,920 \\
Seed               & 42    & 42    \\
\bottomrule
\end{tabular}
\caption{Default inference hyperparameters for each model.
\texttt{gpt-oss-120b} uses greedy decoding; Qwen3-Next uses the recommended sampling parameters from the official documentation.
The max output token limits are discussed in Appendix~\ref{app:threshold}.}
\label{tab:hyperparams}
\end{table}

\subsection{Answer Parsing} 
\label{app:answer_parsing}
To make the answer parsing robust, we prompt the model to reproduce the chosen answer inside an \texttt{<answer>} tag (e.g., \textit{<answer>A) No, they have different underlying distribution</answer>}). We parse the output using edit-distance matching: for each option, we compare the model's response against a length-matched substring of that option, both with and without the leading letter, and assign the nearest match. This handles both outputs that contain only the answer-identifying letter or then ones that omit it.

\subsection{Thinking Output Threshold}
\label{app:threshold}

We limit the maximum number of output reasoning tokens for each model to prevent the model getting stuck in an infinite loop.

\begin{itemize}
  \item For Qwen3-Next, we limit the number of tokens to 81,920 following the official recommendations.\footnote{\url{https://huggingface.co/Qwen/Qwen3-Next-80B-A3B-Thinking-FP8\#best-practices}}
  \item For \texttt{gpt-oss-120b}, we perform preliminary experiments on a \textsc{TimeSeriesExam} dataset, selecting from \{10,20,30,40\},000 tokens; we find 30\,000 tokens to be the best performing threshold. 
\end{itemize}

\subsection{Prompts}
\label{app:prompts}
The full prompts used for each setup are shown in \Cref{fig:prompt_raw,fig:prompt_code,fig:prompt_hybrid}.

\begingroup
\footnotesize
\centering

\begin{prompttextbox}
\begingroup
\ttfamily
\small
\sloppy
\obeyspaces
Given this data:\par
\endgroup

\par\smallskip
\begin{lstlisting}[style=promptjsonstyle]
{data[series]}
\end{lstlisting}

\par\smallskip
\begingroup
\ttfamily
\small
\sloppy
\obeyspaces
Answer the following question by selecting the correct answer and outputting it word-for-word inside of an answer tag without outputting anything else, like this: "<answer>X) lorem ipsum</answer>".\par
\endgroup

\par\smallskip
\begin{lstlisting}[style=promptplainstyle]
{text}
\end{lstlisting}

\end{prompttextbox}
\par
\captionsetup{hypcap=false}
\captionof{figure}{Prompt for the \mdirect{} setup. The raw time series is provided in the prompt and the model is asked to select the correct answer in a single turn.}
\label{fig:prompt_raw}
\endgroup

\begingroup
\footnotesize
\centering

\begin{prompttextbox}
\begingroup
\ttfamily
\small
\sloppy
\obeyspaces
You are given the following question:\par
\endgroup

\par\smallskip
\begin{lstlisting}[style=promptplainstyle]
{text}
\end{lstlisting}

\par\smallskip
\begingroup
\ttfamily
\small
\sloppy
\obeyspaces
Your goal is to answer this question. You have an access to a python code interpreter tool \textasciigrave{}code\textasciigrave{} to gather evidence for your answer. The tool \textasciigrave{}code\textasciigrave{} takes a single argument "code" of type string. You may use the tool multiple times. Gather evidence first before answering. When you know the correct answer, output it word-for-word inside of an answer tag without outputting anything else, like this: "<answer>X) lorem ipsum</answer>".\par
\par\smallskip
When using the code tool, the code should contain a single method \textasciigrave{}main\textasciigrave{} taking a single parameter \textasciigrave{}dict\_of\_dfs\textasciigrave{} of type \textasciigrave{}dict[str, pd.DataFrame]\textasciigrave{}. The argument \textasciigrave{}dict\_of\_dfs\textasciigrave{} is dictionary containing the following key(s): [\{data[json\_keys]\}]. The dataframes under these keys are the time series data. Each dataframe contains a single unnamed column. The python version is 3.9. The libraries available are "pandas==2.2.3", "numpy==1.26.4", "scipy==1.14.1", and "statsmodels==0.14.5".\par
\par\smallskip
Here is an example \textasciigrave{}code\textasciigrave{} start:\par
\par\smallskip
\{"code": "import pandas as pd\textbackslash{}nimport numpy as np\textbackslash{}n\textbackslash{}ndef main(dict\_of\_dfs: dict[str, pd.DataFrame]):\textbackslash{}n    \# \textasciigrave{}dict\_of\_dfs\textasciigrave{} is a dictionary containing the following key(s): [\{data[json\_keys]\}].\textbackslash{}n    \# Under each key is a pandas dataframe with a single unnamed column.\textbackslash{}n    (...)"\}\par
\endgroup

\end{prompttextbox}
\par
\captionsetup{hypcap=false}
\captionof{figure}{Prompt for the \mcode{} setup. The model has no access to raw data and must query the time series via the \texttt{code} tool, which executes a \texttt{main()} function against a \texttt{pandas} DataFrame with the time series data.}
\label{fig:prompt_code}
\endgroup

\begingroup
\footnotesize
\centering

\begin{prompttextbox}
\begingroup
\ttfamily
\small
\sloppy
\obeyspaces
You are given this data:\par
\endgroup

\par\smallskip
\begin{lstlisting}[style=promptjsonstyle]
{data[series]}
\end{lstlisting}

\par\smallskip
\begingroup
\ttfamily
\small
\sloppy
\obeyspaces
And a following question:\par
\endgroup

\par\smallskip
\begin{lstlisting}[style=promptplainstyle]
{text}
\end{lstlisting}

\par\smallskip
\begingroup
\ttfamily
\small
\sloppy
\obeyspaces
Your goal is to answer this question. You have an access to a python code interpreter tool \textasciigrave{}code\textasciigrave{} to gather evidence for your answer. The tool \textasciigrave{}code\textasciigrave{} takes a single argument "code" of type string. You may use the tool multiple times. Gather evidence first before answering. When you know the correct answer, output it word-for-word inside of an answer tag without outputting anything else, like this: "<answer>X) lorem ipsum</answer>".\par
\par\smallskip
When using the code tool, the code should contain a single method \textasciigrave{}main\textasciigrave{} taking a single parameter \textasciigrave{}dict\_of\_dfs\textasciigrave{} of type \textasciigrave{}dict[str, pd.DataFrame]\textasciigrave{}. The argument \textasciigrave{}dict\_of\_dfs\textasciigrave{} is dictionary containing the following key(s): [\{data[json\_keys]\}]. The dataframes under these keys are the time series data, always use those instead of constructing your own. Each dataframe contains a single unnamed column. The python version is 3.9. The libraries available are "pandas==2.2.3", "numpy==1.26.4", "scipy==1.14.1", and "statsmodels==0.14.5".\par
\par\smallskip
Here is an example \textasciigrave{}code\textasciigrave{} start:\par
\par\smallskip
\{"code": "import pandas as pd\textbackslash{}nimport numpy as np\textbackslash{}n\textbackslash{}ndef main(dict\_of\_dfs: dict[str, pd.DataFrame]):\textbackslash{}n    \# \textasciigrave{}dict\_of\_dfs\textasciigrave{} is a dictionary containing the following key(s): [\{data[json\_keys]\}].\textbackslash{}n    \# Under each key is a pandas dataframe with a single unnamed column.\textbackslash{}n    (...)"\}\par
\endgroup

\end{prompttextbox}
\par
\captionsetup{hypcap=false}
\captionof{figure}{Prompt for the \mhybrid{} setup. The raw time series is provided alongside access to the \texttt{code} tool, combining the representations used in \mdirect{} and \mcode{}.}
\label{fig:prompt_hybrid}
\endgroup

\subsection{Judge Prompt}
\label{app:judge_prompt}

The full prompt used for \ljudge{} is shown in \Cref{fig:prompt_judge}. It presents the complete model conversation and the correct answer, then asks the judge to fill in a structured JSON questionnaire following the taxonomy from \Cref{sec:taxonomy}.

\begingroup
\footnotesize
\centering

\begin{prompttextbox}
\begingroup
\ttfamily
\small
\sloppy
\obeyspaces
An LLM agent attempted to answer a question relating to some time series data. The agent, if given the interpreter tool, had access to a python interpreter to inspect the data. Here is the full conversation, including thinking traces (surrounded by <think></think> tags) and tool calls (surrounded by <tool code></tool code> tags):\par
\par\smallskip
=== Agent's conversation ===\par
\par\smallskip
\{data[conv]\}\par
\par\smallskip
=== Conversation end ===\par
\par\smallskip
\par\smallskip
The correct answer was \{data[correct]\}.\par
\par\smallskip
Output a json matching the template below to answer the questions about the agent's conversation. Always explain your decisions under the "X\_expl" field in 1-3 sentences.\par
\endgroup

\par\smallskip
\begin{lstlisting}[style=promptjsonstyle]
{
// "core_reason_for_answer" should be one of the following options:
//  - "code", if the agent answered mainly based on code output.
//  - "raw data", if the agent answered mainly based on "eyeballing" the raw data.
//  - "question format", if the agent answered mainly based on strategizing about the
// exam format.
//  - "other", if none of the options above works.
"core_reason_for_answer":
  enum["code", "raw data", "question format", "other"],
"core_reason_for_answer_expl": string,

// "core_problem_solving_strategy" should be one of the following options, based on the
// primary strategy of the agent:
// - "statistical test", if the primary strategy was to perform a formal hypothesis
// test or compute a test statistic and p-value (examples: Augmented Dickey--Fuller,
// KPSS, Granger causality, t-test, Wilcoxon test, chi-square test, permutation test).
// - "spectral analysis", if the primary strategy was to analyze frequency content of
// the series (examples: FFT, periodogram, Welch's method, Lomb--Scargle, bandpower,
// spectral peak detection).
// - "curve fitting", if the primary strategy was to fit a parametric model to data
// (examples: least-squares polynomial/regression fitting, AR/ARMA/ARIMA/SARIMA
// parameter estimation, nonlinear curve fits, sklearn or statsmodels model.fit usage,
// model-based forecasting).
// - "windowed/rolling statistics", if the primary strategy was to compute statistics
// over sliding or rolling windows, without transforming into the frequency domain
// (examples: rolling mean/variance, rolling correlations or cross-correlations,
// rolling regression, windowed autocorrelation, rolling volatility).
// - "simple arithmetic", if the primary strategy consisted of direct, non-model-based
// summary calculations without statistical inference or signal analysis (examples:
// global mean/median/variance, counts, ratios, totals, expanding statistics, basic
// differencing).
// - "other", if none of the above fits.
"core_strategy":
  enum["statistical test", "spectral analysis", "curve fitting", "windowed/rolling statistics", "simple arithmetic", "other"],
"core_strategy_expl": string,

"methodological_problems": {
  // "conceptual_misunderstanding" should be true if the agent demonstrates a fundamental
  // misunderstanding of key concepts relevant to the question.
  "conceptual_misunderstanding": bool,
  "conceptual_misunderstanding_expl": optional[string],

  // "wrong_core_problem_solving_strategy" should be true if the strategy chosen by the
  // agent ("core_problem_solving_strategy") was not suitable for answering the question.
  // For example, if the model decided to use curve fitting, when using a statistical
  // test would have been better.
  "wrong_core_problem_solving_strategy": bool,
  "wrong_core_problem_solving_strategy_expl":
  optional[string],

  // "wrong_method_within_strategy" should be true if the strategy chosen by the agent
  // ("core_problem_solving_strategy") was suitable for answering the question, but
  // failed due to the specific method chosen within the strategy. For example, the model
  // selected an incorrect statistical test, heuristic, threshold, or algorithmic
  // variant.
  "wrong_method_within_strategy": bool,
  "wrong_method_within_strategy_expl": optional[string],

  // "unsupported_assumption" should be true if for the method that the agent chose, it
  // made an unjustified assumption about the data or the options. For example, assuming
  // stationarity without evidence.
  "unsupported_assumption": bool,
  "unsupported_assumption_expl": optional[string],

  // "implementation_errors" should be true if there were coding or algorithmic mistakes
  // that affect the validity of the result. Code execution errors are that caused the
  // code to fail are not included in this category.
  "implementation_errors": bool,
  "implementation_errors_expl": optional[string],

  // "incorrect_result_interpretation" should be true if the agent incorrectly interprets
  // the result of any of his computations. For example, the model wrongly conclude that
  // the difference is statistically significant.
  "incorrect_result_interpretation": bool,
  "incorrect_result_interpretation_expl": optional[string],

  // "insufficient_evidence_guess" should be true if the agent answered whilst not having
  // gathered sufficiently convincing evidence.
  "insufficient_evidence_guess": bool,
  "insufficient_evidence_guess_expl": optional[string]
},

// This section is primarily meant for agents that had access to a code interpreter
// tool.
"code_problems": {
  // "code_result" should be one of the following options, indicating the overall success
  // of the code the model produced (if any):
  // - "success", if the agent managed to obtain the data it intended. If the code
  // initially failed due to errors, but the agent managed to fix the errors, it still
  // counts as a success.
  // - "partial failure", if the agent failed to obtain some of the desired data, but
  // managed to get at enough results to produce an answer. For example if the code
  // failed after printing sufficient evidence.
  // - "complete failure", if the agent failed to obtain the results and gave up or ran
  // out of turns.
  // - "no code", if the agent didn't produce any code.
  "code_result":
  enum["success", "partial failure", "complete failure", "no code"],
  "code_result_expl": optional[string],

  // "tool_usage_trouble" should be true if the agent had some trouble following
  // instructions to use the tool correctly. This mainly includes cases where the agent
  // generated an improper coding tool call or had trouble calling the tool properly.
  // Regular errors and exceptions that caused the code to fail are not included in this
  // category.
  "tool_usage_trouble": bool,
  "tool_usage_trouble_expl": optional[string],

  // "other" should be true if there was some other issue with the code that does not
  // fall into any of the other categories.
  "other": bool,
  "other_expl": optional[string]
},

"other_problems": {
  // "reasoning_answer_mismatch" should be true if the agent settled on one answer in the
  // thinking trace and then answered differently.
  "reasoning_answer_mismatch": bool,
  "reasoning_answer_mismatch_expl": optional[string],

  // "reasoning_tool_usage_mismatch" should be true if the agent claims it will use the
  // coding tool (such as saying "let's compute", or crafting a code solution in the
  // thinking trace) but then goes straight to answering.
  "reasoning_tool_usage_mismatch": bool,
  "reasoning_tool_usage_mismatch_expl": optional[string],

  // "hallucinated_values_in_reasoning" should be true if the agent introduces numerical
  // values that do not appear in the data or the code output. Reasonable rounding does
  // not count.
  "hallucinated_values_in_reasoning": bool,
  "hallucinated_values_in_reasoning_expl": optional[string]
}
},
\end{lstlisting}

\end{prompttextbox}
\par
\captionsetup{hypcap=false}
\captionof{figure}{Prompt for the LLM judge. The model receives the full conversation trace of the agent, including the question, the reasoning trace, the code (if applicable), and the final answer, along with the correct answer. The model is then asked to fill in a structured JSON questionnaire following the taxonomy described in \Cref{sec:taxonomy}. Note that we abbreviate \texttt{explanation} to \texttt{expl} in the field names for visualization purposes.}
\label{fig:prompt_judge}
\endgroup

\section{Constructing the Error Taxonomy}
\label{app:taxonomy_prompts}

We built an initial set of error categories by running two rounds of Gemini 3 Pro inference on a subset of incorrect \texttt{gpt-oss-120b} outputs. In the first round, Gemini 3 Pro received each incorrect trace paired with the correct answer and produced a short description of where the agent went wrong (Figure~\ref{fig:prompt_error_analysis}). In the second round, we fed all those descriptions into a single prompt asking Gemini to cluster them into fewer than ten categories and assign one to each entry (Figure~\ref{fig:prompt_clustering}).

Gemini returned seven preliminary categories:

\begin{enumerate}
  \item \textbf{Skipping the Coding Phase:} The agent bypassed the coding tool or failed to execute planned code, relying on guesses instead.
  \item \textbf{Statistical Misinterpretation \& Methodological Errors:} Wrong statistical test, misread p-values or $R^2$, or failure to account for stationarity and distribution properties.
  \item \textbf{Visual/Data Hallucination (Manual Inspection):} The agent described values and patterns from raw JSON without computing them.
  \item \textbf{Conceptual \& Semantic Misunderstandings:} Misunderstood core time series concepts (lag, amplitude, variance) or the specific definitions given in the problem.
  \item \textbf{Coding Logic \& Implementation Errors:} Buggy code, wrong library functions, or incorrectly implemented formulas.
  \item \textbf{Flawed Heuristics \& Rigid Thresholds:} Arbitrary numerical thresholds (e.g., $10^{-6}$ tolerance, 0.05 margin) producing incorrect binary decisions.
  \item \textbf{Reasoning--Output Mismatch \& Process Failures:} The agent ignored its own correct reasoning or code output when selecting the final answer.
\end{enumerate}

These categories informed the failure mode section of the final taxonomy (Table~\ref{tab:taxonomy}). We expanded and restructured the section manually, adding the strategy and code problem categories.

\begingroup
\footnotesize
\centering
\begin{prompttextbox}
\begingroup
\ttfamily
\small
\sloppy
\obeyspaces
You are given a conversation with a coding LLM agent. In the conversation, the agent attempts to answer a question by repeatedly producing python code to programmatically query about the relevant time series. The agent answered wrong. The correct answer was "\{data[correct]\}". Analyze where the error happened.\par
\par
Answer in plain text. Do not produce more than 2 paragraphs.\par
\par
=== Agent's Conversation ===\par
\par
\{data[conv]\}\par
\endgroup
\end{prompttextbox}\par
\captionsetup{hypcap=false}
\captionof{figure}{Prompt for per-trace error analysis. Each incorrect model output is analyzed individually.}
\label{fig:prompt_error_analysis}
\endgroup

\begingroup
\footnotesize
\centering
\begin{prompttextbox}
\begingroup
\ttfamily
\small
\sloppy
\obeyspaces
\{analysis 1\}\par
\par
---\par
\par
(...)\par
\par
---\par
\par
\{analysis n-1\}\par
\par
---\par
\par
\{analysis n\}\par
\par
======\par
\par
Each chunk of text represents an analysis of where an LLM coding agent made an error in answering a question about time series. Cluster all these errors into less than 10 error categories. Then output a list assigning each entry the error type. If an entry fits multiple error types, select the most relevant one.\par
\endgroup
\end{prompttextbox}\par
\captionsetup{hypcap=false}
\captionof{figure}{Prompt for clustering the per-trace descriptions into error categories.}
\label{fig:prompt_clustering}
\endgroup

\section{Selecting and Validating \ljudge{}}
\label{app:judge_selection}

\paragraph{Development set.}
We build the development set from the configuration space $\mathcal{C} = \mathcal{D} \times \mathcal{S} \times \mathcal{M}$, where:
\begin{align*}
    \mathcal{D} &= \{\textsc{TimeSeriesExam}, \text{TSFU}\}, \\
    \mathcal{S} &= \{\mmdirect, \mmhybrid, \mmcode\}, \\
    \mathcal{M} &= \{\text{gpt-oss-120b}, \text{Qwen3-Next}\},
\end{align*}
giving $|\mathcal{C}| = 12$. We sample 3 examples per configuration for 36 total. Each configuration's triple contains either 2 correct + 1 incorrect or 1 correct + 2 incorrect answers, chosen with equal probability, keeping the set roughly balanced. Since Qwen3-Next was not evaluated on the full TSFU dataset, the three TSFU $\times$ Qwen3-Next configurations were sampled from a separate pool of 60 randomly selected TSFU questions. No question appears more than once.

\paragraph{Judge configurations.}
We test five judge setups on the development set, all applying the taxonomy from \Cref{sec:taxonomy}:
\begin{itemize}
  \item \textbf{Qwen3-Next (single-phase):} the model generates the required JSON output directly. Constrained decoding prevents it from emitting reasoning tokens, so judgments lack explanations.
  \item \textbf{Qwen3-Next (two-phase):} the model first answers in free-form text, then rewrites its response into the JSON schema in a second turn, recovering reasoning at the cost of an additional inference call.
  \item \textbf{gpt-oss-120b (no system message):} the annotation task is specified entirely in the user message.
  \item \textbf{gpt-oss-120b (system message):} the same model with a custom system message priming it as an expert annotator.
  \item \textbf{gpt-5.4-mini:} an alternative commercial model, helps to better ground the judge capabilities.
\end{itemize}

\paragraph{Results.}
Table~\ref{tab:judge_accuracy} reports agreement with manual labels for each configuration. \texttt{gpt-oss-120b} with a system message scores highest at 90.3\%, while Qwen3-Next single-phase scores lowest at 86.5\%.

\begin{table}[t]
\centering
\begin{tabular}{lc}
\toprule
\textbf{Judge configuration} & \textbf{Accuracy} \\
\midrule
\texttt{qwen3-next-80b} (single-phase)     & 86.5\% \\
\texttt{qwen3-next-80b} (two-phase)        & 88.1\% \\
\texttt{gpt-5.4-mini-2026-03-17}                  & 89.3\% \\
\texttt{gpt-oss-120b} (no system msg.) & 89.3\% \\
\texttt{gpt-oss-120b} (system msg.)    & \textbf{90.3\%} \\
\bottomrule
\end{tabular}
\caption{Agreement of each candidate judge configuration with manual labels on the 36-example development set.}
\label{tab:judge_accuracy}
\end{table}

\begin{table}[t]
    \centering
      \setlength{\tabcolsep}{4pt}
    \begin{tabular}{lrrr}
\toprule
category & \mcode{} & \mdirect{} & \mhybrid{} \\
\midrule
\multicolumn{4}{c}{\texttt{gpt-oss-120b}} \\
\noalign{\vskip 1pt}
\hdashline[0.4pt/2pt]
\noalign{\vskip 1pt}
similarity analysis & 70.0\% & 65.0\% & \textbf{75.8\%} \\
pattern recognition & 75.7\% & 69.1\% & \textbf{80.9\%} \\
causality analysis & 65.3\% & 45.8\% & \textbf{75.0\%} \\
noise understanding & 71.4\% & 76.2\% & \textbf{81.0\%} \\
anolmaly detection & 55.6\% & 57.4\% & \textbf{70.4\%} \\
\midrule
\multicolumn{4}{c}{\texttt{qwen3-next-80b}} \\
\noalign{\vskip 1pt}
\hdashline[0.4pt/2pt]
\noalign{\vskip 1pt}
similarity analysis & 60.6\% & 65.8\% & 65.8\% \\
pattern recognition & 60.2\% & 66.0\% & 70.7\% \\
causality analysis & 55.9\% & 61.1\% & 68.1\% \\
noise understanding & 63.2\% & 67.9\% & 75.0\% \\
anolmaly detection & 35.5\% & 50.9\% & 56.5\% \\
\bottomrule
\end{tabular}

  \caption{Per-category accuracies on \textsc{TimeSeriesExam} for both models and all three setups. The best score for each category across both models and all three setups is in bold.}
    \label{tab:results-full-tse}
\end{table}

\begin{table}[t]
    \centering
          \setlength{\tabcolsep}{4pt}
    \begin{tabular}{lrrr}
\toprule
 & \multicolumn{3}{c}{\texttt{gpt-oss-120b}} \\
\noalign{\vskip 1pt}
\hdashline[0.4pt/2pt]
\noalign{\vskip 1pt}
category & \mcode{} & \mdirect{} & \mhybrid{} \\
\midrule
fixed correlation & \textbf{79.5\%} & 72.2\% & 78.8\% \\
lagged correlation & \textbf{42.5\%} & 30.0\% & 37.5\% \\
anomalies & 65.0\% & 74.0\% & \textbf{75.0\%} \\
seasonality & 63.0\% & 56.0\% & \textbf{72.5\%} \\
fat tail & \textbf{86.5\%} & 69.0\% & 85.0\% \\
stationarity & 46.0\% & 28.0\% & \textbf{51.5\%} \\
structural break & 44.0\% & \textbf{52.0\%} & 49.5\% \\
trend & \textbf{96.5\%} & 88.0\% & \textbf{96.5\%} \\
volatility & 27.5\% & 15.0\% & \textbf{31.0\%} \\
\bottomrule
\end{tabular}

    \caption{Per-category accuracies on \textsc{TSFU} for all three setups. Only \texttt{gpt-oss-120b} results are shown; \texttt{qwen3-next-80b} was not evaluated on the full \textsc{TSFU} dataset (see \Cref{sec:setup}). The best score in each row is in bold.}
    \label{tab:results-full-tsfu}
\end{table}

\clearpage
The accuracy gap between the models is small overall. We select the best performing configuration \texttt{gpt-oss-120b} with a system message as \ljudge{}.

\begin{table*}[b]
  \centering
  \footnotesize
  \setlength{\tabcolsep}{5.2pt}
  \begin{tabular}{@{}lrrrrrrrrrrrrrrrrrrr@{}}
\toprule
 & \multicolumn{3}{c}{\makecell{simple\\arith.}} & \multicolumn{3}{c}{\makecell{stat.\\test}} & \multicolumn{3}{c}{\makecell{rolling\\stat.}} & \multicolumn{3}{c}{\makecell{curve\\fitting}} & \multicolumn{3}{c}{\makecell{spectral\\analysis}} & \multicolumn{3}{c}{other} & \multirow{2}{*}{\textit{n}} \\
\cmidrule(lr){2-4} \cmidrule(lr){5-7} \cmidrule(lr){8-10} \cmidrule(lr){11-13} \cmidrule(lr){14-16} \cmidrule(lr){17-19}
Domain & d & c & h & d & c & h & d & c & h & d & c & h & d & c & h & d & c & h \\
\midrule
\multicolumn{20}{l}{\textit{TimeSeriesExam}} \\[2pt]
\quad Anomaly detection & 67 & 36 & 63 & 0 & 10 & 7 & 0 & 12 & 6 & 0 & 9 & 13 & 0 & 9 & 8 & 33 & 25 & 4 & 638 \\
\quad Pattern recognition & 80 & 36 & 42 & 0 & 18 & 17 & 0 & 3 & 4 & 0 & 24 & 24 & 0 & 10 & 10 & 20 & 8 & 3 & 2,171 \\
\quad Similarity & 81 & 28 & 33 & 0 & 21 & 22 & 0 & 1 & 3 & 0 & 30 & 29 & 0 & 18 & 12 & 19 & 2 & 1 & 718 \\
\quad Noise understanding & 69 & 26 & 32 & 0 & 40 & 40 & 0 & 12 & 10 & 1 & 10 & 10 & 0 & 10 & 8 & 30 & 4 & 1 & 504 \\
\quad Causality analysis & 73 & 2 & 17 & 1 & 29 & 29 & 0 & 68 & 54 & 0 & 1 & 0 & 0 & 0 & 0 & 27 & 0 & 0 & 430 \\
\midrule
\multicolumn{20}{l}{\textit{TSFU}} \\[2pt]
\quad Outliers & 79 & 73 & 95 & 0 & 1 & 2 & 0 & 26 & 3 & 0 & 0 & 0 & 0 & 0 & 0 & 21 & 0 & 0 & 599 \\
\quad Trend & 91 & 0 & 0 & 0 & 78 & 55 & 0 & 0 & 0 & 0 & 22 & 46 & 0 & 0 & 0 & 8 & 0 & 0 & 600 \\
\quad Statistical property & 97 & 93 & 80 & 0 & 7 & 20 & 0 & 0 & 0 & 0 & 0 & 0 & 0 & 0 & 0 & 3 & 0 & 0 & 598 \\
\quad Correlation & 96 & 25 & 15 & 0 & 75 & 85 & 0 & 0 & 0 & 0 & 0 & 0 & 0 & 0 & 0 & 4 & 0 & 0 & 1,198 \\
\quad Lagged correlation & 86 & 3 & 20 & 0 & 83 & 64 & 0 & 14 & 15 & 0 & 0 & 0 & 0 & 0 & 0 & 14 & 0 & 0 & 597 \\
\quad Seasonality & 72 & 16 & 18 & 0 & 0 & 0 & 0 & 0 & 2 & 0 & 0 & 0 & 0 & 84 & 78 & 28 & 0 & 1 & 599 \\
\quad Structural break & 80 & 9 & 2 & 0 & 76 & 88 & 0 & 16 & 9 & 0 & 0 & 0 & 0 & 0 & 0 & 20 & 0 & 0 & 600 \\
\quad Stationarity & 94 & 10 & 7 & 0 & 78 & 81 & 0 & 1 & 3 & 0 & 7 & 6 & 0 & 4 & 2 & 6 & 0 & 0 & 600 \\
\quad Volatility & 82 & 0 & 0 & 0 & 0 & 0 & 2 & 100 & 100 & 0 & 0 & 0 & 0 & 0 & 0 & 16 & 0 & 0 & 600 \\
\bottomrule
\end{tabular}
  \caption{Distribution of core problem-solving strategies per problem domain, split by agent setup (d = \mdirect{}, c~=~\mcode{}, h = \mhybrid{}). Each cell shows the percentage of answers within the given domain and setup, rounded to the nearest integer; within each setup, the dominant strategy is shown in bold. \textit{n} = total number of judge-annotated answers in that domain across all models and setups.}
    \label{tab:domain_strategy}
\vspace{5pt}
\end{table*}

\begin{table*}[b]
\small
\centering
\begin{tabular}{llr rrr r}
\toprule
\textbf{Model} & \textbf{Setup} & \textbf{Data} & \textbf{Mean} & \textbf{Median} & \textbf{Max} & \textbf{Cut off (\%)} \\
\midrule
\texttt{gpt-oss-120b} & \mdirect{} & \textsc{TSE} & 13,340 & 9,933 & 30,040 & 26.8 \\
\texttt{gpt-oss-120b} & \mcode{} & \textsc{TSE} & 4,036 & 2,668 & 30,997 & 4.2 \\
\texttt{gpt-oss-120b} & \mhybrid{} & \textsc{TSE} & 4,693 & 2,459 & 46,622 & 8.2 \\
\texttt{gpt-oss-120b} & \mdirect{} & \textsc{TSFU} & 7,900 & 3,825 & 30,172 & 5.8 \\
\texttt{gpt-oss-120b} & \mcode{} & \textsc{TSFU} & 4,335 & 2,737 & 48,540 & $<$0.1 \\
\texttt{gpt-oss-120b} & \mhybrid{} & \textsc{TSFU} & 3,719 & 2,178 & 43,251 & 0.1 \\
\midrule
\texttt{qwen3-next} & \mdirect{} & \textsc{TSE} & 15,983 & 10,068 & 275,925 & 3.4 \\
\texttt{qwen3-next} & \mcode{} & \textsc{TSE} & 11,886 & 7,496 & 64,730 & $<$0.1 \\
\texttt{qwen3-next} & \mhybrid{} & \textsc{TSE} & 16,730 & 11,219 & 337,976 & 2.5 \\
\bottomrule
\end{tabular}

\caption{Per-question output token statistics by model, setup, and dataset.
Mean and median reflect the typical generation budget per question;
the maximum captures the longest single conversation.
``Cut off'' is the percentage of questions whose answer was explicitly
marked as truncated in the logged outputs.
For \mcode{} and \mhybrid{}, token counts accumulate across all turns
of the multi-turn conversation (code generation, execution feedback,
and the final answer).}
\label{tab:token_usage}
\end{table*}

\newpage
\section{Additional Results}
\label{app:results}
\Cref{tab:results-full-tse,tab:results-full-tsfu} break down accuracy by question category. \Cref{tab:domain_strategy} shows the distribution of strategy choices across domains. \Cref{tab:token_usage} shows the token usage statistics across all setups.


\end{document}